\documentclass{article}
\usepackage{cite}
\usepackage{graphicx} 
\usepackage{algorithm}
\usepackage{algpseudocode}
\usepackage{amsmath}
\usepackage{amssymb}
\usepackage{booktabs}
\usepackage[colorlinks=True,linkcolor=blue,anchorcolor=blue,citecolor=blue,CJKbookmarks=true]{hyperref}

\usepackage{color}

\title{SDEIT: Semantic-Driven Electrical Impedance Tomography}

\begin{document}
\title{SDEIT: Semantic-Driven Electrical Impedance Tomography}
\author{Dong Liu, Yuanchao Wu, Bowen Tong, Jiansong Deng\\
  University of Science and Technology of China, Hefei, China \\
  \texttt{dong.liu@outlook.com} \\
  \texttt{wuyuanchao@mail.ustc.edu.cn} \\
}
\date{}

\maketitle
\newcommand{\enabstractname}{Abstract}
\newenvironment{enabstract}{%
\par\small
\noindent\mbox{}\hfill{\bfseries\enabstractname}\hfill\mbox{}\par
\vskip 2.5ex}{\par\vskip2.5ex}
\newenvironment{cnabstract}{%
	\par\small
	\noindent\mbox{}\hfill{\bfseries\cnabstractname}\hfill\mbox{}\par
	\vskip 2.5ex}{\par\vskip2.5ex}
\begin{enabstract}
Regularization methods using prior knowledge are essential in solving ill-posed inverse problems such as Electrical Impedance Tomography (EIT). However, designing effective regularization and integrating prior information into EIT remains challenging due to the complexity and variability of anatomical structures. 
In this work, we introduce SDEIT, a novel semantic-driven framework that integrates Stable Diffusion 3.5 into EIT, marking the first use of large-scale text-to-image generation models in EIT. SDEIT employs natural language prompts as semantic priors to guide the reconstruction process. By coupling an implicit neural representation (INR) network with a plug-and-play optimization scheme that leverages SD-generated images as generative priors, SDEIT improves structural consistency and recovers fine details. Importantly, this method does not rely on paired training datasets, increasing its adaptability to varied EIT scenarios. Extensive experiments on both simulated and experimental data demonstrate that SDEIT outperforms state-of-the-art techniques, offering superior accuracy and robustness. This work opens a new pathway for integrating multimodal priors into ill-posed inverse problems like EIT.
\noindent{\textbf{keywords:} \quad Electrical Impedance Tomography; \quad Implicit neural representation;\quad Semantic prior;\quad Stable Diffusion Model}
\end{enabstract}
\section{Introduction}
EIT is a promising functional imaging with potential applications in medical diagnosis \cite{gao2024feasibility, kantartzis2013stimulation}, geological exploration \cite{liu2023resolution}, and non-destructive testing \cite{wei2016super}. By measuring surface voltage in response to applied electrical currents, EIT reconstructs the internal conductivity distribution of an object, enabling the detection of structural and functional variations. Its advantages include non-invasiveness, low cost, and real-time imaging capabilities, making it particularly attractive for applications such as lung monitoring, brain imaging, and subsurface exploration. However, despite these benefits, EIT suffers from inherent ill-posedness and low spatial resolution, which significantly limit its widespread adoption. Addressing these challenges requires advanced reconstruction algorithms, effective regularization techniques, and the incorporation of prior information to improve image quality and stability.

Traditional approaches, including Tikhonov regularization \cite{vauhkonen1998tikhonov}, total variation (TV) regularization \cite{gonzalez2016experimental}, and model-based techniques \cite{liu2020shape,liu2020shapeFourier}, have been employed to tackle these challenges. However, these approaches often face difficulties in balancing reconstruction accuracy with computational efficiency, or they require additional prior information.
Meanwhile, leveraging the powerful end-to-end learning capabilities of neural networks (NNs), data-driven reconstruction methods \cite{chen2021deep,hamilton2018deep,seo2019learning} have been introduced in EIT. These methods enhance reconstruction quality by either learning the mapping between voltage measurements and conductivity distributions or refining the reconstructed images through post-processing. Despite their advantages, the inherent complexity of the EIT problem and the limited availability of training data make these methods difficult to train and hinder their generalization performance.

To address these challenges, several self-supervised and unsupervised NNs have been explored in the field of image reconstruction. For instance, the DeepEIT approach \cite{liu2023deepeit} and its network architecture searched version  \cite{nasdip} incorporate Deep Image Prior \cite{ulyanov2018deep} into EIT, while similar architectures have been applied to other image modalities, including CT \cite{baguer2020computed}, PET \cite{gong2018pet}, and MRI \cite{yoo2021time}. DeepEIT effectively generates enhanced reconstructions by optimizing NN parameters without requiring additional training data. However, it struggles to recover high-frequency details due to spectral bias \cite{tancik2020fourier}, where NN tends to prioritize low-frequency components over fine details. To overcome this limitation, implicit neural representation (INR) has been introduced to EIT \cite{wang2023unsupervised}, mitigating spectral bias to some extent while improving convergence and enhancing detail preservation in EIT reconstruction.

Deep generative models (DGMs) have seen rapid development and have shown strong performance in solving image generation problems. 
As the conductivity distribution can be treated as an image, the EIT inverse problem can also be framed as an image generation task. Motivated by this idea, researchers have explored various types of DGMs for EIT reconstruction. For example, classical generative adversarial networks (GANs) have been applied to improve the resolution of EIT images \cite{chen2024mitnet}. However, one inherent limitation of GAN-based methods is the issue of saddle point problems \cite{dauphin2014identifying}.
Recently, diffusion models have emerged as a promising alternative in the field of image generation, offering notable advantages in stability and image quality. These models refine images iteratively through a denoising process, producing high-quality reconstructions across various imaging tasks, and offering an exciting opportunity to overcome some of the challenges associated with GAN-based methods.
Diffusion-based methods in EIT include the conditional diffusion model (CDM) \cite{shi2024conditional,shi2025conditional}, which uses initial reconstruction images or measuring voltages as conditions to incorporate additional information. Alternatively, the Diff-INR method \cite{tong2024diff} introduces a diffusion regularizer and geometric priors to guide the reconstruction, achieving clearer and more accurate reconstructions.

With the development of computing power, multi-modal large language models (MLLMs) have gained significant attention. MLLMs are also widely used in medical imaging. The application fields include classification \cite{tiu2022expert}, medical detection\cite{muller2022radiological}, medical image segmentation\cite{anand2023one} and so on. 
In image reconstruction tasks, including electromagnetic inversion problems, recent studies \cite{zhang2024semantic,chen2024semantic} have leveraged pre-trained multimodal network architectures with large language models (LLMs) like BERT to integrate semantic information into the reconstruction process. 
However, the effectiveness of text comprehension and semantic extraction remains constrained by the characteristics of the dataset.
In contrast, as a representative of MLLMs, stable diffusion models \cite{rombach2022high} integrate the natural language processing capabilities of LLMs with the ability to generate images without requiring additional datasets. Notably, the latest SD3.5 model \cite{esser2024scaling} has demonstrated impressive text-following abilities while maintaining high-quality image generation. This advancement opens new possibilities for applying these models in various fields, including EIT reconstruction.

Inspired by these advancements, we apply SD3.5 to the field of EIT and propose the SDEIT\footnote{\textbf{SD} carries a dual meaning, representing both {\it \textbf{Semantic-Driven}} and {\it \textbf{Stable Diffusion}}, effectively encapsulating the dual role of this framework in integrating semantic information and Stable Diffusion models within the reconstruction process.} framework. In this work, semantic prior knowledge is integrated into the conductivity reconstruction process, enhancing the quality of image generation. 
Our main contributions are as follows:
\begin{itemize}
    \item \textbf{Semantic-driven EIT Reconstruction}: 
    We introduce SDEIT, the first EIT reconstruction method that integrates semantic prior information via Stable Diffusion. 
    The framework optimizes an INR output by iteratively aligning it with SD-generated images through an SSIM-based regularization term, enabling text-conditioned, structure-aware EIT reconstructions.
    \item \textbf{PnP Generative Regularization for EIT}: 
    SDEIT implements a novel Plug-and-Play (PnP) strategy, treating SD as a generative regularizer within the reconstruction loop. This approach enforces structural consistency and enhances the recovery of fine details, overcoming the limitations of conventional regularization schemes and mitigating the spectral bias commonly found in neural network-based EIT methods.
    \item \textbf{Generalizable and Data-Efficient approach}: 
    By eliminating the dependence on paired training data, SDEIT leverages the robust generative priors from SD3.5 to achieve improved generalization across various EIT scenarios. This paired data-free approach enhances adaptability and reduces the need for extensive, scenario-specific data collection.
\end{itemize}

The article is structured as follows: Section \ref{sec.background} provides a brief overview of EIT and INR-based reconstruction methods. Section \ref{sec.SDEIT} introduces the proposed SDEIT framework. The implementation details are outlined in Section \ref{sec.implementaion}. Experimental results are presented in Section \ref{sec.results}, followed by a discussion on challenges and future directions in Section \ref{sec.discussion}. Finally, the conclusions are summarized in Section \ref{sec.conclusion}.

\begin{figure*}[!htb]
\centerline{\includegraphics[width=1\columnwidth]{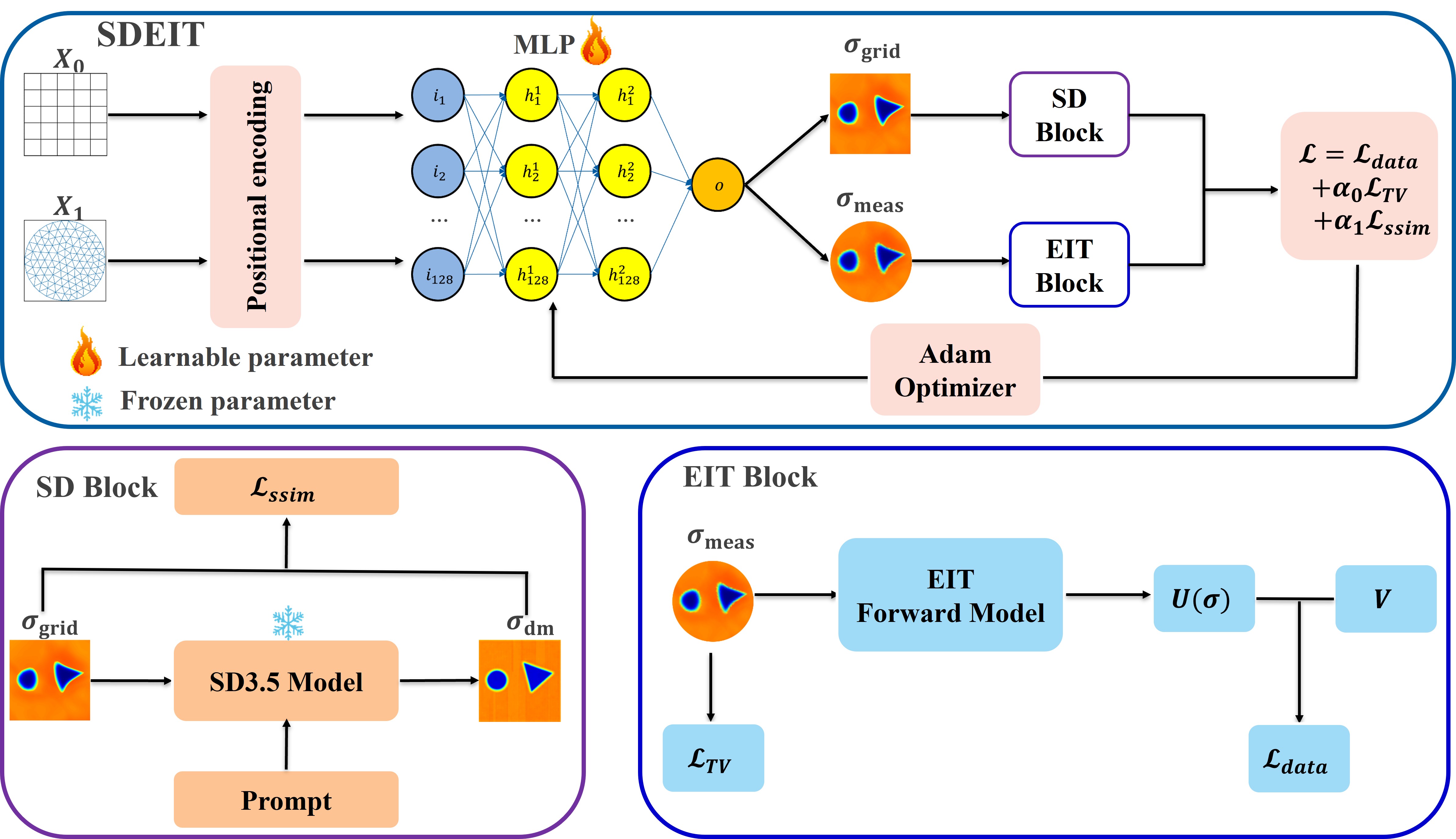}}
\caption{Architecture of the proposed SDEIT framework. The coordinates of FE nodes and the grid are mapped through positional encoding and fed into a four-layer MLP to estimate the conductivity: $\sigma_\text{meas}$ in the measurement domain and $\sigma_\text{grid}$ in the image domain.
$\sigma_\text{meas}$ is processed in the EIT Block to compute the data loss $\mathcal{L}_{data}$ and TV loss $\mathcal{L}_{TV}$, while $\sigma_\text{grid}$ is refined in the SD Block to derive the semantic-based regularization loss $\mathcal{L}_{ssim}$. The model parameters are iteratively updated based on the total loss $\mathcal{L}$ until convergence.}
\label{fig:frame}
\end{figure*}

\section{Background}
\label{sec.background}
In this section, we begin by introducing the EIT forward and inverse problems. We then explore INR-based image reconstruction and its application to solving EIT problems.

\subsection{Forward and inverse problem of EIT}
The EIT forward problem entails computing boundary measurements $V$ from a known conductivity distribution $\sigma$ and injected currents $I$. This relationship is governed by the Complete Electrode Model (CEM) \cite{cheng1989electrode}, which models the behavior of electric fields in a conductive medium. The governing equations are expressed as follows: 

\begin{equation}
    \label{eq:eit}
    \begin{aligned}
\nabla(\sigma \nabla u) & =0, \quad x \in \Omega, \\
u+z_q \sigma \frac{\partial u}{\partial \nu} & =U_q, \quad x \in e_q, \quad q=1, \ldots, N_e, \\
\int_{e_q} \sigma \frac{\partial u}{\partial \nu} \mathrm{d} s & =I_q, \quad q=1, \ldots, N_e, \\
\frac{\partial u}{\partial \nu} & =0, \quad x \in \partial \Omega \backslash \bigcup_{q=1}^{N_e} e_q, \\
\sum_{q=1}^{N_e} I_q & =0, \quad \sum_{q=1}^{N_e} U_q=0, 
\end{aligned}
\end{equation}
where $u$ denotes the electric potential within the domain $\Omega$, $e_q$ represents the position of the $q$-th electrode, and $z_q$ is its corresponding contact impedance. These equations enforce charge conservation and appropriate boundary conditions, ensuring a physically consistent solution.

Given the nonlinearity and 
complexity of the forward model, numerical techniques such as the Finite Element Method \cite{vauhkonen1999three} are widely used to approximate solutions to \eqref{eq:eit}. The forward process can be formulated as an observation model:
\begin{equation}
    \label{eq:observation}
    V=U(\sigma)+e,
\end{equation}  
where $U(\sigma)$ maps the conductivity distribution to the measured voltages, and $e$ represents additive noise. 

Mathematically, the EIT inverse problem can be formulated based on the observation model \eqref{eq:observation} as an optimization problem, where the goal is to minimize the difference between the observed voltage $V$ and the computed voltage $U(\sigma)$.
Due to the ill-posed nature of EIT, direct minimization often results in unstable solutions. 
To overcome this, regularization techniques are usually introduced to impose prior constraints on $\sigma$, leading to the following regularized formulation:  
\begin{equation}
    \label{eq:4}
    \hat{\sigma} = \mathop{\arg\min}\limits_{\sigma} \{\|V - U(\sigma)\|^2 + \alpha R(\sigma)\},
\end{equation}  
where $R(\sigma)$ is a regularization term, and $\alpha$ is a tuning parameter that balances data fidelity and regularization strength. To obtain the solutions with enhanced details, the TV regularization is commonly used, i.e.
\begin{equation}
    \label{eq:loss with TV}
    \hat{\sigma}=\mathop{\arg\min}\limits_{\sigma}\{\|V-U(\sigma)\|^2+\alpha \mathcal{L}_{TV}(\sigma)\}.
\end{equation}
Here, the term TV can be represented as:
\begin{equation}
    \label{eq:TV}
    \operatorname{\mathcal{L}_{TV}}(\sigma)=\sum_{i} \sqrt{\left(D_x \sigma\right)_i^2+\left(D_y \sigma\right)_i^2+\beta},
\end{equation}
where $D_x\sigma$ and $D_y\sigma$ denote the partial derivatives of  conductivity in the $x$ and $y$ directions, and $\beta$ is a small parameter ensuring differentiability.

\subsection{INR-based EIT reconstruction}
As mentioned in the Introduction, INR offers a novel approach to EIT reconstruction. In this method, the (unknown) conductivity distribution is represented as a continuous function parameterized by an NN, denoted by \( \sigma=f(\mathbf{x}; \theta) \), where \( \mathbf{x} \) is a spatial coordinate and \( \theta \) represents the parameters of network $f$. 
To incorporate the spatial information effectively, one usually augments the input coordinates \( \mathbf{x} \) with a positional encoding (PE) function, \( \mathbf{p}(\mathbf{x}) \), which maps the input spatial coordinates to a higher-dimensional space through 
\begin{equation}
\mathbf{p}(\mathbf x)=
\begin{bmatrix}
\sin(2\pi B \mathbf x)\\
\cos(2\pi B \mathbf x)
\end{bmatrix}.
\end{equation}
Here, $B\in \mathbb R^{n\times 2}$ is a matrix of frequencies randomly sampled from the Gaussian distribution $\mathcal N(0,s^2)$. The hyperparameter $s$ determines the bandwidth of the frequencies that the NN can represent. This encoding ensures that the NN can differentiate between distinct spatial positions.

Using INR with PE and TV, the objective of EIT reconstruction is to optimize NN parameters \( \theta \) that minimize the error between the predicted voltage and the observed voltage:
\begin{equation}
    \label{eq:loss with TV (INR)}
    \hat{\theta}=\mathop{\arg\min}\limits_{\theta}\{\|V-U(f(\mathbf{p}(\mathbf{x}); \theta))\|+\alpha_0 \, \mathcal{L}_{TV}(\sigma)\}.
\end{equation}

\section{SDEIT: semantic-driven EIT reconstruction}
\label{sec.SDEIT}
In this section, we present the proposed SDEIT framework, which is structured as two key components for the EIT reconstruction. 
As illustrated in Figure \ref{fig:frame}, the EIT Block focuses on solving the EIT inverse problem, producing a conductivity map constrained by physical principles and enhanced through TV regularization. Concurrently, the SD Block generates semantically enriched guidance images at each iteration, where the SSIM loss is integrated into the optimization process to steer the reconstruction toward outcomes that align with the provided textual descriptions.

Since the measurement domain \( \Omega_{\text{meas}} \) is typically non-rectangular and the SD model operates on pixel-based image domains, we treat the reconstruction area as a rectangular region $[-1,1]\times[-1,1]$ to integrate results from both domains.
The normalized measurement domain is a sub-region within this rectangle, where each point in the image domain or FE node corresponds uniquely to a point in the rectangular region. By feeding its coordinates into MLP, the corresponding conductivity can be obtained. For simplicity, we denote the FE nodes of the measurement domain as $\mathbf{x_0}$ and the grid of the image domain as $\mathbf{x_1}$. Then, the conductivities in the measurement domain and image domain are, respectively, obtained as:  
\begin{equation}
\sigma_\text{meas}=f(\mathbf{p}(\mathbf{x_0});\theta),    
\end{equation}
and
\begin{equation}
\label{sigma_grid}
\sigma_\text{grid}=f(\mathbf{p}(\mathbf{x_1});\theta).    
\end{equation}

\subsection{EIT Block: Physical information alignment}
The EIT Block serves as the foundation for obtaining a physically consistent conductivity by solving the inverse problem.
As shown in Figure \ref{fig:frame}, the EIT Block processes the predicted conductivity $\sigma_\text{meas}$ through the forward model to predict the voltage. The discrepancy between the predicted voltage and the observed voltage is quantified by the data loss $\mathcal{L}_{data}=\|V-U(\sigma)\|^2$, which drives the optimization process. Additionally, the TV term refines the solution, promoting spatial consistency and reducing artifacts, resulting in an initial conductivity estimate aligned with physical principles. This serves as the basis for further refinement in the SD Block, where semantic guidance is introduced to enhance the structural details of the recovered conductivity map.

\subsection{SD Block: incorporating semantic prior}
In this block, the conductivity $\sigma_\text{grid}$, obtained from the MLP-based INR model, serves as an input to the SD Block. 
The SD Block processes $\sigma_\text{grid}$ with a guiding prompt (denoted as $C$) that encodes prior knowledge about the expected structure of the reconstruction. This results in an enhanced conductivity distribution $\sigma_\text{dm}$, which better aligns with the semantic characteristics of the target object.

\subsubsection{Guiding Image Generation with semantic Prior}
To integrate text-based semantic priors into the reconstruction, we begin by encoding the image data and textual prompts into their respective latent representations. Specifically, the Variational Autoencoder (VAE) encoder maps $\sigma_\text{grid}$
to a latent vector, $e_0:\sigma_\text{grid} \mapsto z$, while textual information is encoded through multiple channels: 
CLIP text encoders,$e_1:C\mapsto c_1$, $e_2:C\mapsto c_2$, and the T5 text encoder $e_3:C\mapsto c_3$. 

Next, to introduce image priors into the generation process, 
Gaussian noise is progressively added to the latent vector $z$, resulting in a noisy latent representation $z_{T'}$, 
which is formulated as: 
\begin{equation}
    \label{eq:diff forward}
    q\left( z_t\mid z_{t-1}\right)  =\mathcal{N}\left(z_t ; \sqrt{\alpha_t} z_{t-1},\left(1-\alpha_t\right) I\right) \quad t=0,1, \cdots T',
\end{equation}
\begin{equation}
    \label{eq:diff forward 1}
    z_{T'}  =\sqrt{\overline{\alpha_{T'}}} z_0+\sqrt{1-\overline{\alpha_{T'}}} \epsilon \quad \text { and } \quad z_0=z.
\end{equation}
Here, $T'$ is the forward step, defined as $T'=D\cdot T$, where $D$ represents the denoising strength and $T$ is the diffusion step.

\subsubsection{Denoising and Image Reconstruction}
Starting from \(z_{T'}\), the denoising process iteratively refines the latent representation. At each step \(t\), the transformer integrates the text embeddings \(c = (c_1, c_2, c_3)\) with \(z_t\) and predicts the semantic guided noise component \(\epsilon^{text}_{\theta^{\prime}}(z_t, c, t)\) and unconditional noise component \(\epsilon^{unc}_{\theta^{\prime}}(z_t, t)\), where $\theta^{\prime}$ denotes the parameters of the multimodal transformer. The total noise component is computed as: 
\begin{equation}
    \epsilon_{\theta^{\prime}}=\epsilon^{unc}_{\theta^{\prime}}+G (\epsilon^{text}_{\theta^{\prime}}-\epsilon^{unc}_{\theta^{\prime}}). 
\end{equation}
Here, $G$ denotes the guidance scale. The final output is a latent-space representation of the improved image, denoted as \(z_{\text{lat}}\). The VAE decoder then reconstructs it in the  image domain:  
\begin{equation}
    \sigma_{\text{dm}} = d(z_{\text{lat}}).
\end{equation}  
Then, the entire transformation from \(\sigma_{\text{grid}}\) to \(\sigma_{\text{dm}}\) is denoted as the function \(SD(\cdot)\), and for the \(n\)-th iteration, the guiding image is generated as:  
\begin{equation}
    \sigma_{\text{dm},n} = SD(\sigma_{\text{grid},n}; C, D, T, G).
\end{equation}  

Notably, $\sigma_\text{dm}$ is not directly fed back into the INR model or the EIT Block, except through its role in the loss function. 
This design choice is motivated by the ill-posed nature of EIT, where enforcing a strong prior feedback loop could overly constrain the solution, potentially limiting the INR model’s adaptability to measured data.
Additionally, incorporating $\sigma_\text{dm}$ as a feedback signal would require gradient propagation through the SD Block, leading to significant computational overhead given the scale of modern diffusion models.

\subsubsection{Structural Similarity Loss for Semantic Alignment}
To ensure that the reconstructed image retains both semantic and structural consistency, we employ the mean Structural Similarity Index (mSSIM) as a regularization term. 
The SSIM metric between two image patches is defined as:  
\begin{equation}
    \label{eq:SSIM}
    \operatorname{SSIM}(x, y) = \frac{\left(2 \mu_x \mu_y + K_1\right)\left(2 \sigma_{x y} + K_2\right)}{\left(\mu_x^2 + \mu_y^2 + K_1\right)\left(\sigma_x^2 + \sigma_y^2 + K_2\right)}.
\end{equation}  
Here, $K_1=(k_1 L)^2, K_2=(k_1 L)^2$ are constants ensuring numerical stability, and $L$ denotes the data range. The mean SSIM score is then computed by averaging the SSIM values over all image patches:  
\begin{equation}
    \label{eq:mSSIM}
    \operatorname{mSSIM}(X, Y) = \frac{1}{M} \sum_{i=1}^{M} SSIM(x_i, y_i).
\end{equation}  
Here, \(M\) denotes the total number of patches. To enforce alignment between the reconstructed image and the semantic prior, we define the structural loss function as:  
\begin{equation}
    \mathcal{L}_{ssim}(\sigma_{\text{grid},n}, \sigma_{\text{dm},n}) = 1 - \operatorname{mSSIM}(\sigma_{\text{grid},n}, \sigma_{\text{dm},n}).
\end{equation}  

\subsubsection{Total Loss and Optimization Process}
The complete loss function for the \(n\)-th iteration integrates data fidelity, TV regularization, and structural similarity loss:  
\begin{equation}
    \label{eq:loss with SSIM}
    \begin{aligned}
     \mathcal{L}_n(\theta) = \|U - V(\sigma_{\text{meas},n})\|^2 + &\alpha_0 \mathcal{L}_{TV}(\sigma_{\text{meas},n}) \\  
     &+ \alpha_1 \mathcal{L}_{ssim}(\sigma_{\text{grid},n}, \sigma_{\text{dm},n}).
    \end{aligned}
\end{equation}  
Since each term in Equation \eqref{eq:loss with SSIM} is differentiable, the NN parameters \(\theta\) are updated through backpropagation. 

After performing the full iterations, the final reconstruction is obtained, ensuring alignment with both the physical constraints  from the EIT data and the semantic prior from the SD block. The entire procedure is outlined in Algorithm \ref{alg:Framwork}.

\begin{algorithm}[!htb]
\caption{SDEIT}
\label{alg:Framwork}
\begin{algorithmic}[1]
\Require
  Coordinates: $\mathbf{x_0}$ and $\mathbf{x_1}$;
  INR parameters $\theta$;
  Epochs $N_0$ and $N$;
  Learning rate $lr$;
  Measured voltages $V$;
  Prompt $C$;
  Denoising strength $D$;
  Diffusion step $T$;
  Guidance scale $G$;
  Regularization weights $\alpha_0$, $\alpha_1$.
  
\State \textbf{for} $e=0$ to $N-1$ \textbf{do}: 
\State \qquad \textit{\# Map coordinates to feature space}
\State \qquad$\mathbf{x_0^{\prime}}=\mathbf{p}(\mathbf{x_0})$, $\mathbf{x_1^{\prime}}=\mathbf{p}(\mathbf{x_1})$ 
\State \qquad \textit{\# Obtain conductivity distribution from INR}
\State \qquad$\sigma_\text{meas}= f(\mathbf{x_0^{\prime}};\theta)$, $\sigma_\text{grid}= f(\mathbf{x_1^{\prime}};\theta)$ 
\State \qquad Compute predicted voltage: $U=U(\sigma_\text{meas})$

\State \qquad \textbf{if} $e<N_0$ \textbf{then} \textit{\# Early-stage optimization}
\State \qquad \quad $\mathcal{L}=\|U-V\|^2+\alpha_0\mathcal{L}_{TV}$
\State \qquad \quad Compute gradient: 
\[
\frac{\partial \mathcal{L}}{\partial \theta}=2\left(\frac{\partial U}{\partial \sigma_{\text {meas }}}\right)^T\left(\frac{\partial \sigma_{\text {meas}}}{\partial \theta}\right)(U-V)+\alpha_0\frac{\partial \mathcal{L}_{TV}}{\partial \theta}
\]
\State \qquad \textbf{else} \textbf{\# Semantic-guided refinement begins}
\State \qquad \quad Generate refined conductivity: 
\[\sigma_\text{dm}=SD(\sigma_{\text{grid}};C,D,T,G)\]
\State \qquad \quad Compute structural loss: 
\[\mathcal{L}_\text{ssim}=\mathcal{L}_\text{ssim}(\sigma_\text{grid},\sigma_\text{dm})\]
\State \qquad \quad Define total loss:
\[
\mathcal{L}=\|U-V\|^2+\alpha_0\mathcal{L}_{TV}+\alpha_1\mathcal{L}_{ssim}
\]
\State \qquad \quad Compute gradient:
\[
\begin{aligned}
 \frac{\partial \mathcal{L}}{\partial \theta} &= 2\left(\frac{\partial U}{\partial \sigma_{\text {meas }}}\right)^T 
\left(\frac{\partial \sigma_{\text {meas}}}{\partial \theta}\right)(U-V) \\
& \quad + \alpha_0\frac{\partial \mathcal{L}_{TV}}{\partial \theta} 
+ \alpha_1\frac{\partial \mathcal{L}_\text{ssim}}{\partial \theta}
\end{aligned}
\]
\State \qquad \textbf{end if}
\State \qquad \textit{\# Update INR parameters using Adam optimizer}
\State \qquad Update $\theta$ with Adam optimizer
\State \textbf{end for}

\Ensure
 $\sigma_\text{grid}$, $\sigma_\text{meas}$, $\sigma_\text{dm}$
\end{algorithmic}
\end{algorithm}

\section {Implementation details}
\label{sec.implementaion}
In this section, we begin by outlining the setup for both simulation and experimental studies, followed by a presentation of the evaluation metrics used and a detailed description of the parameter settings applied in the test cases.

\subsection{Experiments Setup}
\label{sec:Experiments Setup}
In the numerical simulations, we evaluate two distinct test cases. The first, a simple yet representative case, involves a disk with a 14 cm radius as the measurement domain. Sixteen electrodes, each with a width of 2.5 cm, are placed equidistantly along the boundary. Electric currents with an amplitude of 1 mA are injected into the measurement domain, with current stimulation and voltage measurement following adjacent patterns. The tissue conductivities are set as 0.25 mS/cm for the ellipse-shaped lungs, 1 mS/cm for the background, and 1.5 mS/cm for the circular-shaped heart. Gaussian noise with an SNR of 60 dB is added to simulate real-world conditions. The mesh used for the forward problem consists of $N_n = 5833$ nodes and $N_e = 11424$ elements, while the inverse mesh contains $N_n = 1145$ nodes and $N_e = 2176$ elements.

The second test case is based on a thorax-shaped domain derived from a CT scan of a human thorax. On the domain's boundary, 16 electrodes, each 2 cm in length, are placed equidistantly for use in EIT measurements. The conductivity of the background is set to 2 mS/cm, while the heart is assigned 3 mS/cm and the lungs 0.8 mS/cm. To test the robustness of the proposed SDEIT framework, noise levels ranging from 60 to 40 dB are added to the noiseless data. The mesh for the forward problem consists of $N_n = 7150$ nodes and $N_e = 13802$ elements, and the inverse mesh contains $N_n = 5586$ nodes and $N_e = 10674$ elements.

For the experimental test cases, we used a water tank with a 14 cm radius, filled with saline solution, as the measurement domain. The tank was equipped with 16 metallic rectangular electrodes, each 7 cm in height, evenly distributed around its periphery. Four non-conductive objects of varying shapes were placed within the tank as reconstruction targets. The experimental data were collected using the KIT-4 measurement system \cite{kourunen2008suitability}. The meshes used for this setup are identical to those employed in the first simulated heart-and-lungs phantom.
It is important to note that all meshes used in this study are first-order meshes. Distinct mesh designs are employed for solving the forward and inverse problems to avoid the occurrence of the {\it inverse crime} phenomenon.

\subsection{Evaluation Metrics}
To quantify the imaging quality of SDEIT, we use two metrics: the mSSIM as defined in \eqref{eq:mSSIM} and the Peak Signal-to-Noise Ratio (PSNR), given by:
\begin{equation}
    \label{eq:PSNR}
     \operatorname{PSNR}=10 \cdot log_{10}(\frac{\text{MAX}_\text{I}^2}{\text{MSE}}).
\end{equation}
Here, $\text{MAX}_\text{I}$ is the maximum possible value, and Mean-Square Error (MSE) is defined as follows: 
\begin{equation}
    \label{eq:MSE}
    \operatorname{MSE} = \frac{1}{n} \sum_{i=1}^{n} (\sigma_i - \sigma_i^{\prime})^2,
\end{equation}
where $\sigma_i$ and $\sigma_i^{\prime}$ represent the values of the reconstructed conductivity and the ground truth, respectively.

The mSSIM index is determined using a sliding window approach, with a window size of 7 applied throughout the process. 
For the reconstruction results from experimental data, in addition to these two metrics, we evaluate the proximity and correlation to the real conductivity distribution by calculating MSE and the Correlation Coefficient (CC):
\begin{equation}
    \label{eq:CC}
     \operatorname{CC}=\frac{\sum\left(\sigma_i-\bar{\sigma}\right)\left(\sigma^{\prime}_i-\bar{\sigma}^{\prime}\right)}{\sqrt{\sum\left(\sigma_i-\bar{\sigma}\right)^2 \sum\left(\sigma^{\prime}_i-\bar{\sigma}^{\prime}\right)^2}} ,
\end{equation}
where $\bar{\sigma}$ and $\bar{\sigma}^{\prime}$ denote the mean value of the reconstructed conductivity and the ground truth.

\subsection{Execution and Optimization}
\label{sec:Execution and Optimization}
For the representation of the conductivity distribution, we selected a four-layer MLP with 128 neurons per layer. The coordinates are encoded using a sampling count of $n = 128$, which is then provided as input to the MLP. 
The stable diffusion model employed in our work is SD 3.5 (stable-diffusion-3.5-large-turbo), except for the preliminary study on fine-tuning in Section \ref{sec.discussion}, where we explored the potential for adapting the model to specific tasks.
The model parameters are remaining fixed throughout the process.
In our experiments, we first use basic semantic guidance to generate the guiding images. The basic semantics are described through simple prompts, such as 
\begin{itemize}
    \item \it basic prompt= (``Shapes. Clean background. Simple form."),
\end{itemize}
ensuring that the image generation process focuses on structural information without being influenced by complex semantics. 
When exploring the impact of additional semantic information, we introduce more detailed and specific textual descriptions, which provide higher-level semantic guidance. In this way, the model not only focuses on the structural information of the image but also incorporates more complex semantic insights, leading to a deeper influence on the reconstruction task. 

The other input parameters of the SD3.5 model are set as $(D,T,G)=(0.4,50,0.8)$. This parameter configuration ensures that during the generation of guidance images, the structural information of the input image plays a dominant role, while semantic guidance serves as a supplementary factor. As a result, the model's output aligns more closely with the physical information, enhancing the accuracy and reliability of the reconstructed images.

The regularization weights are set to $\alpha_0 = 10^{-6}$ and $\alpha_1 = 10^{-2}$ for simulated data, while for experimental data, $\alpha_1 = 3 \times 10^{-3}$ is used. For the additional simulations conducted in the discussion of Section \ref{sec.discussion}, we adjust the regularization weights to $\alpha_0 = 10^{-4}$ and $\alpha_1 = 10^{-2}$ to ensure the stability of the iterative process. The number of iterations is set to $N_0 = 800$ for the pre-guidance phase and $N = 1200$ for the full iteration process. The Adam optimizer's learning rate is set to 0.01.
It is important to note that these parameters are empirically chosen through trial and error.
All experiments are conducted on a system running Ubuntu 22.04 with Python 3.10.12, featuring an AMD EPYC 7763 CPU and an Nvidia GTX 4090D 24GB GPU.

\section{Results}
\label{sec.results}
In this section, we provide the results obtained using both simulated and experimental data.
For simplicity, we denote the conventional reconstruction using TV regularization as $\sigma_{\text{TV}}$, the reconstruction with the recently developed INR using TV regularization as $\sigma_{\text{INR+TV}}$, and the reconstruction based on the proposed SDEIT framework as $\sigma_{\text{SD}}$. 
Since the primary focus of this work is to incorporate semantic priors into EIT reconstruction, operating as a paired data-free reconstruction framework, we do not include comparison studies against paired data-driven supervised reconstructions.

\subsection{Simulation Results}
\begin{figure}[!htb]
\centerline{\includegraphics[width=\columnwidth]{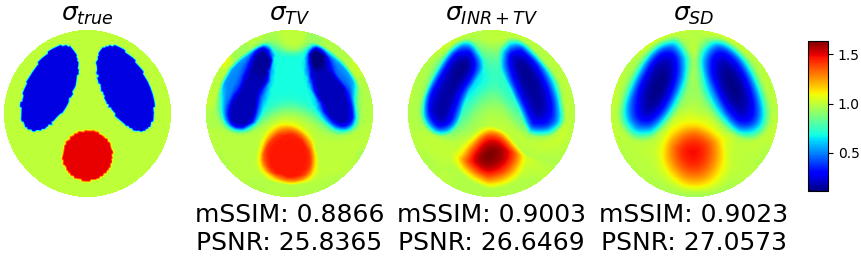}}
\caption{Simulation results comparing EIT reconstructions using TV,  INR+TV, and the proposed SDEIT method.}
\label{fig:simu31}
\end{figure}

The simulation study presented in Figure \ref{fig:simu31} highlights the comparative performance of the traditional TV-based and INR+TV-based EIT reconstruction methods. The TV-regularized solution provides basic structural recovery but struggles with over-smoothing, leading to blurred inclusion boundaries and reduced quantitative scores (mSSIM = 0.8866, PSNR = 25.84). By integrating INR, the  INR+TV method enhances detail preservation, yielding improved mSSIM (0.9003) and PSNR (26.65) values. However, minor artifacts and slight edge smoothing persist.
In contrast, the proposed SDEIT method outperforms both baselines, achieving the highest mSSIM (0.9023) and PSNR (27.06). Leveraging semantic priors provided by the {\it basic prompt} enables SDEIT to guide the reconstruction towards sharper, more structure-consistent solutions while preserving fine details and reducing noise.
These results demonstrate SDEIT's ability to surpass conventional and hybrid methods by incorporating text-conditioned generative regularization, providing a promising new avenue for robust and high-fidelity EIT imaging.

\begin{figure}[!htb]
\centerline{\includegraphics[width=\columnwidth]{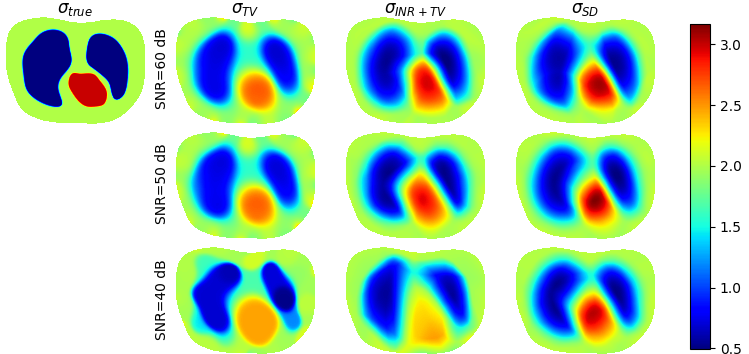}}
\caption{Robustness analysis of TV, INR+TV, SDEIT using a simulated thorax phantom under varying noise levels.}
\label{fig:thorax_full}
\end{figure}

Figure \ref{fig:thorax_full} presents the results of the robustness study conducted on a human thorax phantom under varying SNR conditions (60 dB, 50 dB, and 40 dB). Across all noise levels, SDEIT-based $\sigma_{\text{SD}}$ consistently outperforms other methods by better preserving structural details and maintaining sharper region boundaries. As the SNR decreases, traditional TV-based reconstructions ($\sigma_{\text{TV}}$) suffer from increased blurring and artifacts, while the INR-based method with TV regularization ($\sigma_{\text{SD+TV}}$) improves upon $\sigma_{\text{TV}}$ by recovering more fine-grained structures but still exhibits excessive smoothing. In contrast, SDEIT demonstrates superior robustness to noise, producing conductivity distributions that closely resemble the ground truth. 
This is further supported by the evaluation criteria shown in Table \ref{tab:robustness}, where mSSIM decreases only slightly from 0.8775 to 0.8722, and PSNR drops marginally from 24.79 to 24.07 as the SNR decreases from 60 dB to 40 dB, highlighting SDEIT’s noise tolerance. These results emphasize the effectiveness of incorporating semantic priors via the SD Block, particularly in maintaining anatomical fidelity under low-SNR conditions.

Considering that EIT is inherently affected by noise and modeling uncertainties, these findings suggest that SDEIT is well-suited for real-world clinical and industrial applications where noise resilience is essential. 
Additionally, reconstructing the heart remains challenging due to EIT’s lower sensitivity in central regions and the shielding effect of  the surrounding lungs. Despite these inherent limitations, SDEIT leverages semantic priors to enhance reconstruction quality, demonstrating its potential for improving EIT reconstructions in anatomically complex scenarios.

\begin{table}[!htb]
 \centering
 \caption{Performance comparison of the proposed SDEIT approach with the reference methods in various noise levels.}
  \resizebox{1\columnwidth}{!}{
      \begin{tabular}{ccccccc}
        \toprule
        \multicolumn{1}{c}{\textbf{}} & \multicolumn{2}{c}{\textbf{SNR=60 dB}} & \multicolumn{2}{c}{\textbf{SNR=50 dB}} & \multicolumn{2}{c}{\textbf{SNR=40 dB}}  \\  
        \cmidrule(lr){2-3} \cmidrule(lr){4-5} \cmidrule(lr){6-7}
        \textbf{} & \textbf{mSSIM} & \textbf{PSNR} & \textbf{mSSIM} & \textbf{PSNR} &\textbf{mSSIM}  & \textbf{PSNR}  \\
        \midrule
        \textbf{SDEIT} & \textbf{0.8775} & \textbf{24.79} & \textbf{0.8728} & \textbf{24.18} & \textbf{0.8722} & \textbf{24.07} \\
        \textbf{INR+TV} & 0.8756 & 24.41 & 0.8722 & 23.88 & 0.8713 & 23.40 \\
        \textbf{TV} & 0.8727 & 24.19 & 0.8706 & 24.05 & 0.8697 & 23.76 \\
        \bottomrule
      \end{tabular}
      }
  \label{tab:robustness}
\end{table}

\subsection{Experimental Results}
\begin{figure}[!htb]
\centerline{\includegraphics[width=\columnwidth]{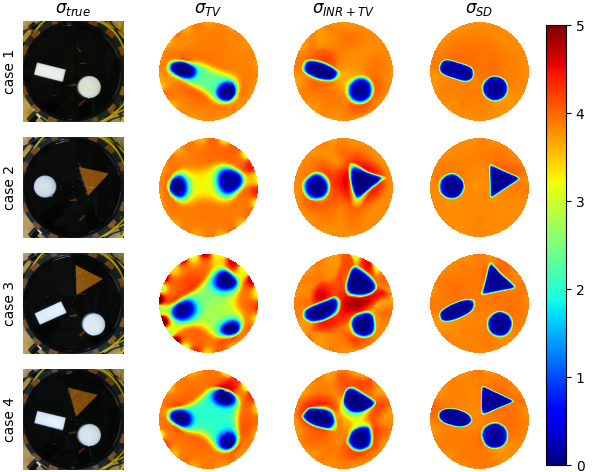}}
\caption{Experimental phantom reconstructions using TV, INR+TV, and the proposed SDEIT method across four test cases. Each row corresponds to a different experimental setup, showing the ground truth (left) and reconstruction results (right three columns).}
\label{fig:aku}
\end{figure}

Figure \ref{fig:aku} displays the experimental validation results for the SDEIT method compared with conventional TV-based and INR+TV-based reconstructions. The conventional TV-based approach provides reasonably accurate localization of targets but is hindered by severe artifacts and poorly defined object boundaries, largely due to the staircase effect. INR+TV mitigates these artifacts, producing clearer object shapes and a cleaner background, though some structural inaccuracies persist. 

In contrast, SDEIT demonstrates superior performance, achieving sharper object boundaries and minimal background noise. The circular and triangular objects are accurately reconstructed, resembling the ground truth. While the rectangular object is accurately localized, its geometry is approximated as an elliptical shape, highlighting the challenge of precisely capturing sharp corners when the guiding prompt encodes only general prior information.
Further comparisons between reconstructions guided by {\it basic} and {\it full prompts} will be discussed in the following subsection. Overall, SDEIT outperforms both baseline methods in preserving structural integrity and suppressing artifacts. The quantitative results in Table \ref{tab:aku} further support SDEIT’s superior performance.

\begin{figure}[!htb]
\centerline{\includegraphics[width=\columnwidth]{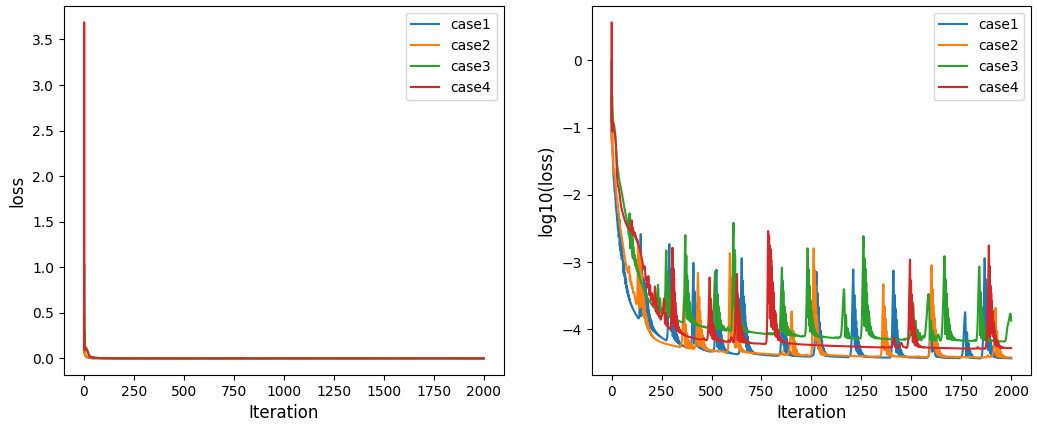}}
\caption{Loss curve (left) and log loss (right) plot of the proposed SDEIT approach against iteration steps for experimental test cases 1-4.}
\label{fig:SD_loss}
\end{figure}

As outlined in Section \ref{sec:Execution and Optimization}, the reconstruction was completed after $N=1200$ iterations. To further evaluate the performance of the proposed method beyond this point and facilitate a fair comparison with the INR+TV reconstruction (which is terminated at 2000 iterations), we extended the iteration steps for SDEIT to $N=2000$, matching the INR+TV setup. Meanwhile, to illustrate the rapid error reduction and stable reconstruction with semantic guidance in the proposed method, we plotted the loss curve and corresponding log loss curve in Figure \ref{fig:SD_loss}. 

SDEIT demonstrates rapid convergence, significantly reducing the data error within the initial $\sim200$ iterations. The periodic fluctuations observed in the log-scale loss curve after approximately 250 iterations are indicative of the model's efforts to refine the conductivity distribution, leading to subtle adjustments. Despite these oscillations, the log loss values remain consistently low, fluctuating between -2 and -5, which is acceptable given the inherently noisy nature of EIT inverse problems. Such variability is characteristic of the reconstruction process as the model focuses on enhancing finer structural details after capturing the global anatomy. Notably, following the introduction of semantic guidance at iteration $N_0=800$, the data error stabilizes further, with the overall loss trend maintaining a similar trajectory to the pre-guidance phase. This suggests that the  semantic priors enhance structural consistency while preserving adherence to the physical constraints inherent in the EIT framework.

\begin{table*}[!htb]
 \centering
 \caption{Performance comparison of the proposed SDEIT approach with the reference methods.}
  \resizebox{1\columnwidth}{!}{
      \begin{tabular}{ccccccccccccccccc}
        \toprule
        \multicolumn{1}{c}{\textbf{}} & \multicolumn{4}{c}{\textbf{Case 1}} & \multicolumn{4}{c}{\textbf{Case 2}} & \multicolumn{4}{c}{\textbf{Case 3}} & \multicolumn{4}{c}{\textbf{Case 4}} \\  
        \cmidrule(lr){2-5} \cmidrule(lr){6-9} \cmidrule(lr){10-13} \cmidrule(lr){14-17}
        \textbf{} & \textbf{mSSIM} & \textbf{CC} & \textbf{PSNR} & \textbf{MSE} & \textbf{mSSIM} & \textbf{CC} & \textbf{PSNR} & \textbf{MSE} &\textbf{mSSIM} & \textbf{CC} & \textbf{PSNR} & \textbf{MSE} &\textbf{mSSIM} & \textbf{CC} & \textbf{PSNR} & \textbf{MSE} \\
        \midrule
        \textbf{SDEIT} & \textbf{0.930} & \textbf{0.88} & \textbf{24.47} & \textbf{0.022} & \textbf{0.928} & \textbf{0.86} & \textbf{23.39} & \textbf{0.029} & \textbf{0.880} & \textbf{0.78} & \textbf{20.52} & \textbf{0.069} & \textbf{0.876} & \textbf{0.76} & \textbf{20.25} & \textbf{0.069}\\
        \textbf{INR+TV} & 0.924 & 0.86 & 23.22 & 0.030 & 0.914 & 0.83 & 21.05 & 0.050 & 0.867 & 0.77 & 18.70 & 0.084 & 0.865 & 0.74 & 19.25 & 0.074\\
        \textbf{TV} & 0.911 & 0.81 & 22.09 & 0.039 & 0.899 & 0.82 & 21.00 & 0.049 & 0.784 & 0.70 & 16.35 & 0.145 & 0.808 & 0.68 & 17.54 & 0.110\\
        \bottomrule
      \end{tabular}
      }
  \label{tab:aku}
\end{table*}


\subsection{Exploring the Influence of Semantic Prior}
\begin{figure}[!htb]
\centerline{\includegraphics[width=.82\columnwidth]{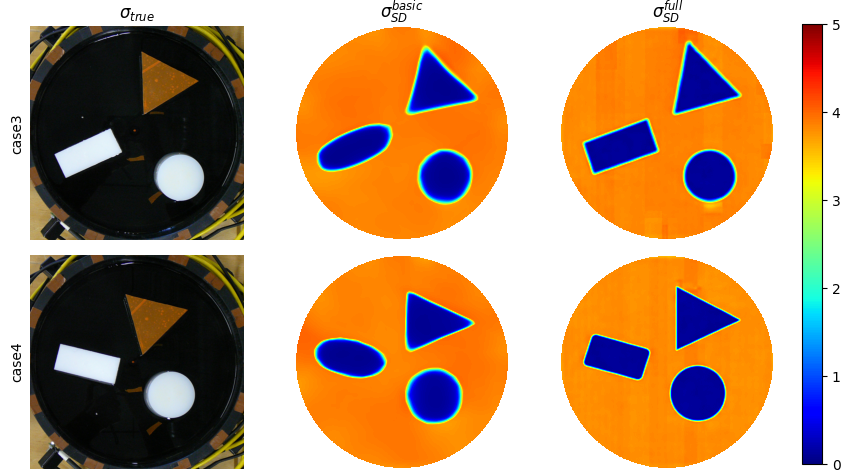}}
\caption{Experimental results investigating the impact of semantic priors on SDEIT reconstructions for case 3 and case 4.}
\label{fig:aku_full}
\end{figure}

In previous experiments, only the basic prompt was used to evaluate the performance of the SDEIT framework. 
However, the integration of various semantic priors plays a crucial role in the SDEIT framework.
To further examine the influence of semantic information, we conducted additional experiments on case 3 and case 4, employing both {\it basic prompt} (as introduced in Section \ref{sec:Execution and Optimization}) and an extended {\it full prompt} defined as:
\begin{itemize} 
\item \it full prompt = (``Shapes. The one at the upper right is a triangle. The one at the lower left is a rectangle. Clean background. Simple form."). 
\end{itemize}
The results, illustrated in Figure \ref{fig:aku_full}, reveal that incorporating full semantic descriptions significantly enhances shape reconstruction accuracy, especially for rectangular objects, which display sharper edges and more clearly defined corners. Since all other parameters remained unchanged, these improvements can be primarily attributed to variations in the guidance images generated with the SD Block of the proposed SDEIT framework.

\begin{figure}[!htb]
\centerline{\includegraphics[width=\columnwidth]{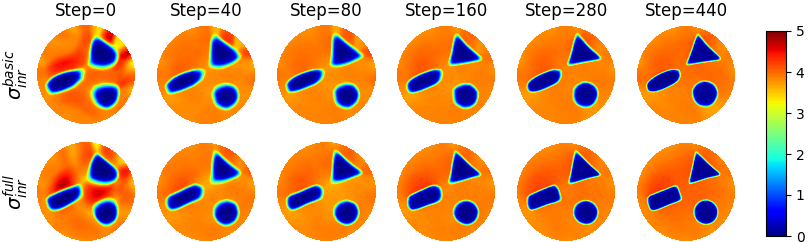}}
\caption{Comparison of the reconstructed images with the proposed SDEIT approach using {\it basic} and {\it full} semantic prompts at different guiding steps.}
\label{fig:sd_iter}
\end{figure}

\begin{figure}[!htb]
\centerline{\includegraphics[width=\columnwidth]{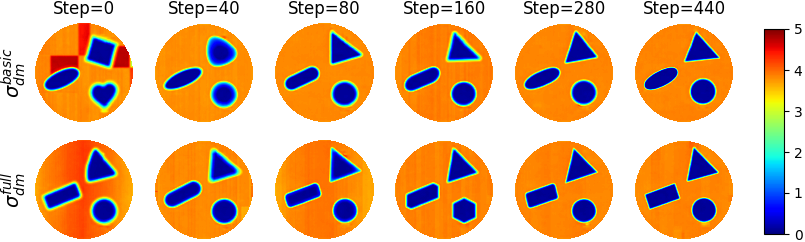}}
\caption{The guiding images of the proposed SDEIT approach with {\it basic} and {\it full} semantic prompts at different guiding steps.}
\label{fig:guiding image}
\end{figure}

To illustrate how semantic priors influence the process, Figures \ref{fig:sd_iter} and \ref{fig:guiding image} show the progression of SDEIT reconstructions over various iteration steps in the post-guidance phase, alongside the corresponding guidance images generated by the SD3.5 model. 
Note that the iteration steps $i$ in the post-guidance phase of Figures \ref{fig:sd_iter} and \ref{fig:guiding image} correspond to the iteration steps $N_0+i$ of the entire reconstruction process, as we incorporate the semantic priors starting at step $N_0=800$. 
Additionally, since the reconstructions after $i=440$ steps are nearly identical, we omit the remaining reconstructions and guiding images for clarity.

It is evident that, under basic semantic prompts, early-stage guidance images often contain vague or distorted shapes. While these images gradually improve and stabilize as iterations proceed, they still fail to fully capture the rectangular geometry. Conversely, when full semantic prompts are provided, the SD3.5 model generates highly precise guidance images, enabling the SDEIT method to achieve more accurate and reliable reconstructions, especially for rectangular targets.
This outcome is expected. In the context of the EIT reconstruction problem, when local conductivity distributions remain relatively consistent, small variations in object boundaries may have a minimal impact on the data error $\mathcal{L}_{data}$. As a result, the optimization process focuses on reducing the SSIM loss $\mathcal{L}_{ssim}$ by aligning the reconstructed image with the spatial distribution suggested by the guidance image, while maintaining stable conductivity values. 
Given that these guidance images incorporate the provided semantic information, this mechanism effectively integrates semantic priors into the reconstruction process.

\section{Discussion}
\label{sec.discussion}
Despite the significant advancements achieved by the current SDEIT framework, several challenges remain for its broader application in medical imaging.
First, human organs exhibit complex, irregular, and highly variable geometries, which make them fundamentally different from the simple geometric shapes—such as circles, triangles, or rectangles—commonly used in basic simulation studies. 
To enhance SDEIT’s applicability to real medical scenarios, it is essential to move beyond basic prompts and adopt medically and anatomically informed terminology when incorporating prompts for generative models like Stable Diffusion.
For thorax EIT, one could employ composite prompts that incorporate shape, position, and texture, such as:
\begin{itemize} 
\item \it Prompt = (``An EIT image showing an irregular, lobed heart surrounded by two asymmetrically shaped lungs with soft, smooth textures and realistic organ borders."). 
\end{itemize}
However, despite using highly detailed prompts, we found that SDEIT exhibited limitations in producing accurate reconstructions. 
This shortcoming is likely attributed to the lack of EIT-specific images within the pretraining data of Stable Diffusion, constraining its ability to represent such distributions effectively. 
\begin{figure}[!htb]
\centerline{\includegraphics[width=.9\columnwidth]{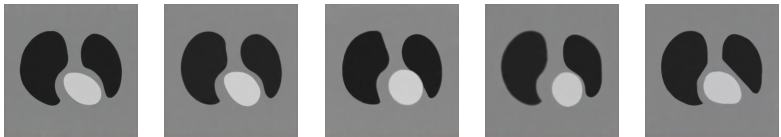}}
\caption{The domain-specific images used to fine-tune the Stable Diffusion model with LoRA.}
\label{fig:train_pic}
\end{figure}

\begin{figure}[!htb]
\centerline{\includegraphics[width=1\columnwidth]{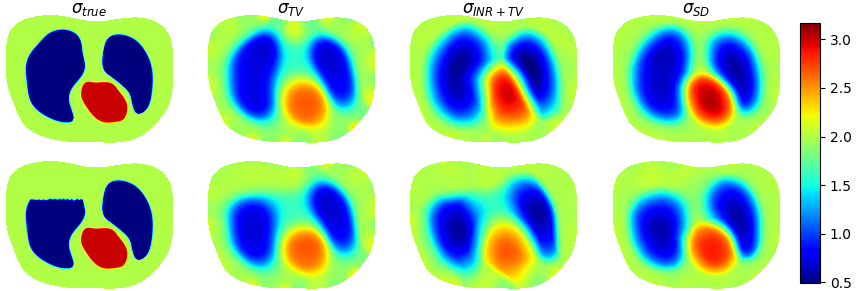}}
\caption{Thorax EIT reconstruction results comparing TV,  INR+TV, and the proposed SDEIT framework utilizing fine-tuned semantic prompts (e.g., \textit{$<$EiThx$>$}) for structural guidance.}
\label{fig:EiThx}
\end{figure}

To address this, lightweight fine-tuning methods \cite{han2024parameter} can be employed as a potential solution by teaching the model new, domain-specific concepts tailored to the EIT reconstruction task. Techniques such as Textual Inversion \cite{gal2022image}, LoRA (Low-Rank Adaptation) \cite{hu2022lora}, and DreamBooth \cite{ruiz2023dreambooth} enable the model to internalize anatomical structures and imaging patterns not present in its original training samples of Stable Diffusion. Textual Inversion allows to introduce new pseudo-tokens representing complex anatomical regions, thereby enabling the generation of more accurate and concise medical guidance images. LoRA and DreamBooth, on the other hand, fine-tune specific components or layers of the model using limited domain-specific datasets, allowing Stable Diffusion to adapt to specialized imaging modalities. 
In our initial experiments, we utilized five grayscale images resembling thoracic anatomy (Figure \ref{fig:train_pic}) to fine-tune the SD3.5-medium model with LoRA, embedding the learned distribution as a pseudo-token (e.g., \textit{$<$EiThx$>$}). 
We tested this fine-tuned model under two scenarios: one using a standard lung-and-heart phantom and another using a phantom with a lung partially removed. The corresponding prompt: \begin{itemize} \item \it Prompt = (``An image of EiThx. Simple form."). \end{itemize} 
As shown in Figure \ref{fig:EiThx}, the first row highlights that when the target structure closely resembles the fine-tuning dataset, SDEIT achieves precise shape reconstruction. In contrast, the second row demonstrates that even when the target deviates from the training images, the model still offers adaptable structural guidance, indicating that the fine-tuned distribution retains a notable level of generalization and transferability.

These preliminary findings suggest that adapting generative models through fine-tuning can enable the creation of anatomically accurate and clinically relevant guidance images.
This, in turn, enhances the accuracy and realism of semantic-driven  reconstructions, especially in complex inverse problems such as  EIT, where the recovery of internal structures is fundamentally challenging. 
While the primary focus of this paper is the development of the SDEIT framework, we recognize that systematically incorporating these fine-tuning techniques represents a promising direction for future research in advancing medical imaging technologies.

\section{Conclusion}
\label{sec.conclusion}
In this paper, we introduced SDEIT, a novel semantic-driven framework for EIT reconstruction. 
Unlike conventional EIT reconstruction methods, which are often constrained by ill-posedness and limited structural information, 
SDEIT integrates INR with semantic priors derived from a frozen Stable Diffusion 3.5 model. By reparameterizing the conductivity as the output of a coordinate-based NN and optimizing its parameters, SDEIT effectively refines the reconstruction results.

Our framework is entirely training-free and does not require additional datasets. Instead, it leverages semantically enriched guidance images generated by Stable Diffusion to steer the INR-based optimization process toward more realistic and artifact-suppressed reconstructions.
Through extensive numerical simulations and experimental validations, we demonstrated that SDEIT outperforms traditional approaches, such as TV-based reconstruction, as well as the recently developed INR+TV method, in terms of both visual quality and quantitative metrics. Furthermore, we studied the role of semantic priors, revealing that richer and more detailed prompts contribute significantly to reconstruction accuracy, especially when dealing with complex conductivity properties.

While the current study focuses on basic and canonical shapes for proof-of-concept, we identified limitations when transitioning to real-world medical imaging scenarios involving anatomically complex organs. To address this, we discussed potential extensions involving lightweight fine-tuning techniques (e.g., Textual Inversion, LoRA, DreamBooth) to adapt generative models like Stable Diffusion to domain-specific contexts such as thorax EIT.

Overall, SDEIT opens new directions for integrating modern generative models and semantic priors into medical imaging and inverse problems. Future work will extend this framework to clinical datasets, explore advanced prompt engineering for medical applications, and incorporate domain-adapted diffusion models to enhance reconstruction performance in real-world settings.

\bibliographystyle{unsrt}
\bibliography{reference}
\end{document}